%% file: root.tex
\documentclass[journal]{IEEEtran}

\usepackage[ruled, lined, linesnumbered, commentsnumbered, longend]{algorithm2e}

\usepackage{amsfonts}
\usepackage{amssymb}
\usepackage{bm}
\usepackage{amssymb}
\usepackage{algpseudocode}
\usepackage{array}
\usepackage{textcomp}
\usepackage{gensymb}
\usepackage{pstricks}
\usepackage{cite}
\usepackage{color}
\usepackage{graphicx}
\usepackage[outdir=./]{epstopdf}
\graphicspath{{./pdf/}}
\usepackage[cmex10]{amsmath} % \usepackage{amsfonts}
\usepackage{amsfonts}
\usepackage{amssymb}
\usepackage{amsmath}
\usepackage{bm}
\usepackage{amssymb}
\usepackage{algpseudocode}
\usepackage{array}
\usepackage{textcomp}
\usepackage{gensymb}
\usepackage{pstricks}
\usepackage{hyperref}
\hypersetup{hidelinks}
\usepackage{CJKutf8}
\usepackage{pifont}

\usepackage{booktabs}
\usepackage[skip=3pt]{caption} % for figure caption skip setting
\usepackage[skip=1pt,labelformat=empty]{subcaption} % for subfigure caption skip setting, and remove label format
\usepackage{siunitx}
\usepackage{float}
\usepackage{nccmath}
\usepackage{xcolor,soul}
\usepackage{changes}
\usepackage{verbatim}
\usepackage{enumerate}
\usepackage{amsmath}
\usepackage{multirow}

\usepackage{amsthm}
\theoremstyle{definition}

\usepackage{xcolor}

\usepackage{color}
\usepackage{soul}
\sethlcolor{yellow}

\newif\ifshowchanges
\showchangesfalse
\newcommand{\rev}[1]{%
  \ifshowchanges
    \hl{#1}%
  \else
    #1%
  \fi
}

\soulregister{\cite}{7}
\soulregister{\ref}{7}

\begin{document}
%
% paper title
% Titles are generally capitalized except for words such as a, an, and, as,
% at, but, by, for, in, nor, of, on, or, the, to and up, which are usually
% not capitalized unless they are the first or last word of the title.
% Linebreaks \\ can be used within to get better formatting as desired.
% Do not put math or special symbols in the title.
\title{MILE: A Mechanically Isomorphic Hand Exoskeleton and Visuotactile Robotic Hand for Data Collection in Dexterous Manipulation}

\author{Jinda~Du$^{*1,2,3}$,
Jieji~Ren$^{*1,2}$,
Qiaojun~Yu$^{3}$,
Ningbin~Zhang$^{1,2}$,
Yu~Deng$^{4}$,
Xingyu~Wei$^{1}$,
Yufei~Liu$^{4}$,
Guoying~Gu$^{\dagger1,2}$,
and Xiangyang~Zhu$^{\dagger1,2}$

\thanks{This work was supported in part by the National Key R\&D Program of China under Grant No.~2024YFB4707504; in part by the National Natural Science Foundation of China under Grant Nos.~52305029 and 52505029; and in part by the Science and Technology Commission of Shanghai Municipality (STCSM) under Grant Nos.~25ZR1401191 and 24511103401.}

\thanks{1. State Key Laboratory of Mechanical System and Vibration, School of Mechanical Engineering, Shanghai Jiao Tong University, Shanghai 200240, China.}

\thanks{2. Shanghai Key Laboratory of Intelligent Robotics, Shanghai Jiao Tong University, Shanghai 200240, China.}

\thanks{3. Shanghai Artificial Intelligence Laboratory, Shanghai, China.}

\thanks{4. Humanoid Robot (Shanghai) Co., Ltd., Shanghai, China.}

\thanks{$^*$ These authors contributed equally to this work. }
\thanks{$^\dagger$ Corresponding authors (mexyzhu@sjtu.edu.cn, guguoying@sjtu.edu.cn).}
}

\maketitle

\begin{abstract}
Dexterous robotic hands perform complex, contact-rich manipulation. Imitation learning provides a route to such skills, but collecting human demonstrations with accurate hand actions and rich tactile information remains a key bottleneck. We present MILE, a teleoperation-based data-collection system comprising the wearable MILE exoskeleton and the mechanically corresponding MILE-Tac robotic hand. Because human-hand anatomy and wearability place tighter constraints on the high-DoF wearable, our human-first design begins with the MILE exoskeleton, equipped with custom modular joint encoders for accurate joint-angle acquisition. We then design the MILE-Tac robotic hand to share the exoskeleton's selected kinematic topology and joint-axis arrangement while satisfying robot-side implementation constraints, and equip its fingertips with compact visuotactile sensor modules. This correspondence enables direct exoskeleton-to-robot joint-space command transfer without online task-space inverse-kinematics retargeting. During teleoperation, the system synchronously records task-specific visual observations, four fingertip visuotactile streams, robot-hand proprioception, and exoskeleton-derived action commands. In a four-task teleoperation benchmark, MILE achieved a mean success rate of 76\%, compared with 28\% and 8\% for glove-based and vision-based baselines, respectively. For downstream imitation learning, we trained paired ACT and DP policies with and without tactile input on MILE-collected demonstrations. The tactile-input variants achieved higher success rates in all paired evaluations.

\end{abstract}


\section{Introduction}

\begin{figure}[!t]
    \centering
    \includegraphics{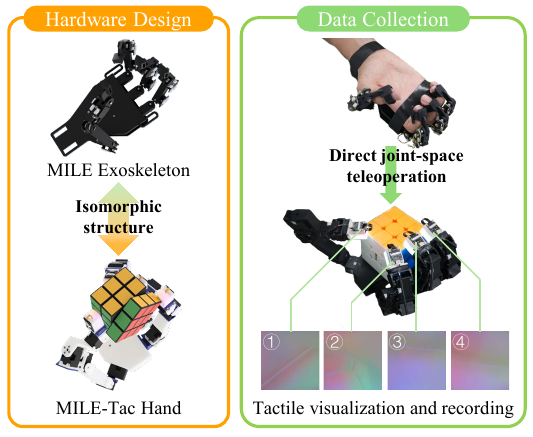}
    \caption{Overview of the \textbf{MILE data-collection system}. The
system comprises two main components: the MILE exoskeleton with integrated joint encoders and the mechanically
corresponding MILE-Tac robotic hand with four integrated fingertip visuotactile sensor modules.
The exoskeleton provides joint-space commands for teleoperation, while the
robotic hand executes the corresponding motions. Fingertip tactile images are
visualized to indicate contact states during teleoperation and are recorded
together with mapped robot-hand joint commands and robotic-hand joint states.}
    \label{fig:teaser}
\end{figure}

Embodied artificial intelligence is pushing robotic manipulation
toward greater dexterity, expanding end-effectors from simple parallel
grippers~\cite{chiUniversalManipulationInterface2024a} to high-DoF
dexterous robotic hands~\cite{andrychowicz2020learning,zhang2025biomimetic}
capable of contact-rich in-hand manipulation. The high dimensionality
and complex contact dynamics of such hands make coordinated manipulation
challenging for model-based control. Learning-based methods therefore
provide a complementary data-driven route. Among them, reinforcement
learning~\cite{pmlr-v80-haarnoja18b} can learn policies for dexterous
contact-rich manipulation tasks, but it typically requires task-specific
reward design~\cite{wang2024penspin}, can suffer from sim-to-real
gaps~\cite{tobin2017domain,chenVisualDexterityInhand2023b}, and may
remain sample-inefficient in high-dimensional exploration spaces. In
contrast, imitation learning leverages real-robot demonstrations~\cite{kim2021transformer}
and provides a practical approach to learning policies for contact-rich
manipulation tasks~\cite{liu2022robot}.

However, imitation learning~\cite{chi2025diffusion,zhao2023learning}
requires large volumes of human-demonstrated dexterous-manipulation
trajectories~\cite{wu2024gello}. Common
demonstration-collection interfaces for dexterous robotic hands include
markerless vision-based hand-pose
estimation~\cite{huangLearningDexterityHuman2024,cheng2024tv},
marker-based optical motion capture (mocap), and wearable instrumented
interfaces, including glove-based
systems~\cite{wangDexCapScalablePortable2024,yinDexterityGenFoundationController2025}
and articulated hand
exoskeletons~\cite{zhangDOGloveDexterousManipulation2025,fang2025dexop,
benHOMIEHumanoidLocoManipulation2025}.
When the kinematics represented by the operator-side interface do
not match those of the target robotic hand, task-space retargeting or
workspace projection may be
required~\cite{meattiniHumanRobotHand2023,benHOMIEHumanoidLocoManipulation2025}.

Contact-rich dexterous manipulation presents perceptual challenges
because hand--object contacts are frequently occluded from external vision,
while tactile sensing provides complementary contact information that is
unavailable from vision alone~\cite{ye2026visualtactile}. Optical
visuotactile sensors can further capture high-resolution local contact
geometry and sensing-surface deformation~\cite{yuan2017gelsight}. These local contact observations provide task-relevant information for contact-rich dexterous manipulation that is difficult to recover from external vision alone. We therefore integrate robot-side fingertip visuotactile sensing into MILE.

As shown in Fig.~\ref{fig:teaser}, we introduce \textbf{MILE}
(\textbf{M}echanically \textbf{I}somorphic \textbf{L}inkage
\textbf{E}xoskeleton), a human-first, mechanically isomorphic design
paradigm for teleoperation and data collection in contact-rich dexterous
manipulation. In this study, human-first denotes a constraint-prioritization strategy
in which human-hand anatomy and wearability requirements guide the exoskeleton
design, while robot-side realizability constrains the implementation of the
MILE-Tac hand. Mechanical isomorphism in MILE refers to the correspondence
between the exoskeleton and the MILE-Tac hand.

Consistent with this human-first strategy, the MILE exoskeleton is an anthropometrically informed wearable system for operator-side joint-command acquisition. The MILE-Tac hand is subsequently designed to share the exoskeleton’s kinematic topology and joint-axis arrangement, while its dimensions and packaging are adapted to accommodate robot-side actuators and fingertip sensors. It further integrates four compact, modular fingertip visuotactile sensors that provide high-resolution robot-side tactile observations. At the implementation level, the
MILE-Tac hand adopts the modular and low-cost design principles of the LEAP
Hand family for actuator packaging, fabrication, and
maintenance~\cite{shawLEAPHandLowCost2023}.

MILE establishes a hardware-level kinematic correspondence across 17 joint
coordinates, including two-axis MCP joints together with PIP and DIP joints
for the non-thumb fingers, and two-axis CMC and MCP joints together with an
IP joint for the thumb. Within this selected topology, the measured exoskeleton joint angles
correspond directly to the commanded robotic-hand joint angles.
This joint-space correspondence enables direct transfer of corresponding
joint configurations, maintaining kinematic consistency between the
exoskeleton and robotic hand without the distortions or feasibility issues
that may arise from task-space retargeting, while eliminating the need for
online task-space IK-based retargeting in the exoskeleton-to-robot command
mapping.

During teleoperation, the exoskeleton provides operator-side joint commands,
while the robotic-hand tactile streams can be displayed as auxiliary visual
contact information. We evaluate MILE against representative glove-based and
vision-based teleoperation interfaces and compare policies trained with and
without fingertip tactile input using Action Chunking Transformer (ACT) and
Diffusion Policy (DP) backbones.
We further conduct two additional experiments using the MILE-Tac hand
mounted on a robotic arm: fixed-bottle cap unscrewing and sequential
potato-chip pick-and-place.

\textbf{Our main contributions are summarized as follows:}
\begin{itemize}
\item A human-first, constraint-driven hardware design strategy that
establishes a 17-DoF exoskeleton--robot joint-space correspondence without
online task-space IK-based retargeting.

\item A synchronized multimodal demonstration-collection platform that integrates accurate exoskeleton joint sensing and fingertip visuotactile sensing for contact-rich dexterous manipulation.

\item System-level experimental validation across teleoperation and
autonomous dexterous manipulation tasks using MILE-collected visual--tactile
demonstrations.

\end{itemize}

The project page, including design files for the fingertip
visuotactile sensor, is available at
\url{https://jindadu00.github.io/MILE}.

\section{Related Work}

\subsection{Manipulation Data Collection for Dexterous Robotic Hands}

Markerless vision-based interfaces recover operator hand motion directly from
camera observations and enable immersive teleoperation, but the accuracy and
temporal robustness of the estimated human-hand pose can be degraded by
self-occlusion, scene clutter, and tracking failures~\cite{cheng2024tv}. The resulting pose errors may then propagate
through downstream retargeting to the commanded robotic-hand motion.
Marker-based optical systems can provide accurate three-dimensional tracking
but typically depend on markers, camera infrastructure, and a calibrated
workspace.

Wearable instrumented interfaces include glove-based systems and
articulated exoskeletons. Glove-based interfaces use compliant textile or finger-mounted supports carrying motion sensors, such as flex, inertial, or electromagnetic sensors, to capture human-hand motion for teleoperation and demonstration collection~\cite{wangDexCapScalablePortable2024,yinDexterityGenFoundationController2025,saggio2025emfglove}.
However, their performance can depend on fit, variation in hand size and
shape, sensor placement, calibration, and embodiment mapping. Articulated exoskeletons use stiff links and explicit joints mechanically coupled to the hand, providing mechanically constrained joint sensing for dexterous teleoperation and manipulation~\cite{zhangDOGloveDexterousManipulation2025,
fang2025dexop,benHOMIEHumanoidLocoManipulation2025}. However, the wearable
mechanism must fit around the hand while providing sufficient clearance around
the finger joints and between neighboring fingers, and may increase wearer
encumbrance. When the kinematics represented by the operator-side interface do not match
those of the target robotic hand, interfaces in any of these categories may
require task-space retargeting or workspace projection, which may involve
scaling mismatch, singularity handling, or command clipping
~\cite{meattiniHumanRobotHand2023}.

A complementary strategy is to reduce retargeting through hardware-level
kinematic correspondence between the operator-side interface and the target
robot. Although developed for robot-arm teleoperation, GELLO uses a scaled,
kinematically equivalent controller whose joints and links correspond directly
to those of the target arm, allowing measured joint positions to be mapped
directly to target-arm joint commands~\cite{wu2024gello}.
Extending this principle to wearable dexterous-hand interfaces is more challenging because the operator-side mechanism must satisfy human-hand anatomical, fit, and clearance constraints across multiple joint axes.
DexUMI reduces this kinematic mismatch through target-hand-driven
hardware adaptation, optimizing a wearable exoskeleton for a target
robotic hand to match its fingertip kinematics, with flexibility in the
remaining linkage geometry to satisfy wearability constraints~\cite{xu2025dexumi}.
In contrast, MILE adopts a human-first, constraint-driven design in which
the selected wearable 17-DoF topology and joint-axis arrangement serve as
the basis for the exoskeleton--robot hardware correspondence, while the
MILE-Tac geometry is adapted to robot-side actuation and sensor packaging.
A more detailed comparison is provided in Supplementary Section~II.

\subsection{Tactile Data Collection and Policy Learning for Dexterous Manipulation}

Many existing imitation-learning systems for dexterous manipulation rely primarily on vision and proprioception, which provide limited direct observability of distributed contact interactions. Tactile sensing has therefore been explored through multiple modalities with different sensing characteristics and integration trade-offs~\cite{luo2025tactileroboticsoutlook}. Beyond transduction modality, fingertip-surface compliance is also an important design variable because it affects contact conformity, load distribution, and tactile sensitivity. Excessive softness can reduce spatial or temporal fidelity, so the appropriate compliance is task- and sensor-dependent~\cite{luo2025tactileroboticsoutlook}. Force/torque sensors provide global wrench measurements but do not directly resolve spatially distributed contact over the fingertip surface~\cite{luo2025tactileroboticsoutlook}. Electronic skins provide compact, distributed sensing but generally require sensor-specific calibration and signal conditioning~\cite{geBimodalSoftElectronic2019,luo2025tactileroboticsoutlook}. Optical visuotactile sensors can provide high-resolution contact geometry and shear-related cues~\cite{yuan2017gelsight,ncpalm}. Optical visuotactile sensing also entails practical trade-offs: deformable sensing surfaces can wear or be damaged, camera and illumination hardware can increase fingertip volume and mass, and dense image acquisition and processing impose bandwidth and computational costs that may limit the system-level frame rate. Many representative tactile-learning datasets and methods have focused on two-finger grippers~\cite{yuan2025tactile,huang3d} and have demonstrated benefits on contact-rich manipulation tasks~\cite{10505922,huang2025vtrefine,zhao2025polytouchrobustmultimodaltactile}. For multi-fingered dexterous hands, high-resolution tactile integration remains less common, reflecting challenges in compact multi-fingertip sensor integration and the availability of suitable multimodal datasets. Recent work has demonstrated visual–tactile pretraining and online multitask learning for multifingered manipulation using monocular visual observations and binary tactile signals~\cite{ye2026visualtactile}. Complementary to these efforts, MILE integrates a compact visuotactile
sensor module at each fingertip of the robotic hand for synchronized tactile
recording during teleoperated demonstration collection. The collected tactile
observations are further evaluated in downstream imitation-learning
experiments.

\begin{figure}[!t]
	\centering
	\includegraphics[width=\linewidth]{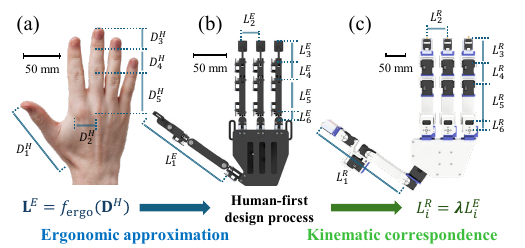}
\caption{Human-first hardware realization of the MILE
exoskeleton--hand system.
(a) Representative human-hand dimensions $\mathbf{D}^{H}$ guiding the
wearable-side design.
(b) MILE exoskeleton designed from the representative human-hand dimensions
under ergonomic and wearability constraints.
(c) MILE-Tac hand designed to share the selected four-finger topology and
corresponding joint-axis arrangement with the exoskeleton, with the
corresponding structural dimensions uniformly scaled by the factor
$\lambda=9/5$ to accommodate robot-side actuator packaging.}
	\label{fig:Isomorphic_comparison}
\end{figure}

\section{System Design}

\begin{figure*}[t]
	\centering
	\includegraphics[width=\linewidth]{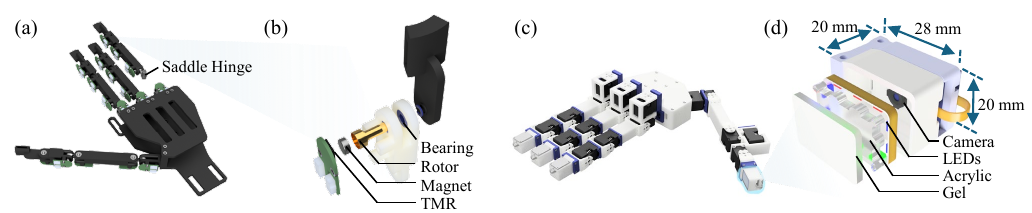}
\caption{Hardware views of the MILE system and key components.
(a) MILE exoskeleton with a 5-DoF thumb and 4-DoF index, middle, and ring fingers.
(b) Representative joint module with an integrated joint-angle encoder.
(c) MILE-Tac hand with integrated fingertip visuotactile sensor modules.
(d) Detailed view of the fingertip visuotactile sensor module.}
	\label{fig:exploded_view}
	\vspace{-0.6\baselineskip}
\end{figure*}

We propose a mechanically isomorphic system for synchronized multimodal
demonstration collection in dexterous manipulation.
Fig.~\ref{fig:Isomorphic_comparison} summarizes the hardware
realization of the human-first design. Let $\mathbf{D}^{H}$ denote the
representative human-hand dimensions shown in
Fig.~\ref{fig:Isomorphic_comparison}(a), and let $\mathbf{L}^{E}$ denote the
selected structural dimensions of the MILE exoskeleton. The notation
$\mathbf{L}^{E}=f_{\mathrm{ergo}}(\mathbf{D}^{H})$ schematically denotes
the ergonomic design mapping from human-hand dimensions to the wearable
geometry. The MILE-Tac hand realizes the selected topology and corresponding
joint-axis arrangement. For the implemented exoskeleton-to-robot design,
we adopt a geometric scale factor $\lambda$ as a practical implementation
choice. The robotic-hand geometry, including corresponding link dimensions,
joint-axis locations, and non-intersecting joint-axis offsets, is uniformly
scaled to provide packaging space for the actuators while maintaining the
selected joint-axis arrangement.

\subsection{Design Criteria}
The MILE exoskeleton--hand system is designed according to the following criteria:

\begin{itemize}
\item \textbf{Isomorphism}: The MILE exoskeleton and the MILE-Tac hand
share the selected kinematic-tree topology and establish a calibrated affine
correspondence between measured exoskeleton joint coordinates and commanded
robotic-hand joint coordinates, enabling direct joint-space command transfer
without online task-space IK-based retargeting.

\item \textbf{Wearability}: Following the human-first design principle,
the exoskeleton geometry and adjustable attachments are informed by human-hand
anatomy and wearability considerations to support user-friendly teleoperation
while preserving the intended finger motions.
\item \textbf{Precision}: Accurate and repeatable exoskeleton joint-angle acquisition.
\item \textbf{Modularity}: Reconfigurable joint modules and standardized interfaces that facilitate fabrication, maintenance, and future kinematic reconfiguration.
\end{itemize}

Specifically, within the selected 17-DoF exoskeleton--robot design, we define the exoskeleton--robot correspondence as follows. Let
$\mathcal{G}_h=(\mathcal{V}_h,\mathcal{E}_h)$ and
$\mathcal{G}_r=(\mathcal{V}_r,\mathcal{E}_r)$ denote rooted kinematic
trees of the exoskeleton and robotic hand, respectively, and let
$\mathbf{q}_h,\mathbf{q}_r\in\mathbb{R}^{n}$ denote their joint
coordinates. The joint-edge mapping
$\pi:\mathcal{E}_h\rightarrow\mathcal{E}_r$ is a selected
topology-preserving one-to-one mapping that preserves parent--child incidence and
finger-chain ordering. It induces the index-permutation matrix
$\mathbf{P}_{\pi}$, with the convention
$[\mathbf{P}_{\pi}\mathbf{q}_h]_j=q_{h,\pi^{-1}(j)}$.

All geometric quantities below are evaluated in the calibrated nominal
configuration. The vector $\mathbf{a}_{\cdot,j}$ denotes the
unoriented unit direction of joint axis $j$; $\mathbf{p}_{\cdot,j}$
denotes a CAD-defined reference point on joint axis $j$; and
$\ell_{\cdot,j}$ denotes the corresponding downstream axis-to-axis or
nominal link dimension. The rotation $R\in\mathrm{SO}(3)$
and translation $\mathbf{t}\in\mathbb{R}^3$ align the selected
exoskeleton and robotic-hand base frames. The scale factor
$\lambda>0$ maps exoskeleton dimensions to robotic-hand dimensions,
with $\lambda=9/5$ for the implemented system. The intervals
$[q^{\min}_{h,j},q^{\max}_{h,j}]$ and
$[q^{\min}_{r,j},q^{\max}_{r,j}]$ denote the mechanically limited
exoskeleton joint range and the robotic-hand mechanical joint range,
respectively.

The two mechanisms satisfy the selected mechanically isomorphic
correspondence within tolerances
$(\boldsymbol{\alpha},\boldsymbol{\varepsilon}_{p},
\boldsymbol{\varepsilon}_{\ell})$ if the following four constraints hold for
all corresponding joints:
\begin{align}
\arccos\!\left(
\left|
\mathbf{a}_{r,j}^{T}R
\mathbf{a}_{h,\pi^{-1}(j)}
\right|
\right)
&\leq \alpha_j,
\qquad \forall j,
\label{eq:axis}\\
\left\|
\mathbf{p}_{r,j}
-\lambda R\mathbf{p}_{h,\pi^{-1}(j)}
-\mathbf{t}
\right\|
&\leq \varepsilon_{p,j},
\qquad \forall j,
\label{eq:axis_location}\\
\left|
\ell_{r,j}
-\lambda\ell_{h,\pi^{-1}(j)}
\right|
&\leq\varepsilon_{\ell,j},
\qquad \forall j,
\label{eq:length}
\end{align}
\begin{equation}
\begin{aligned}
&\bigl\{
s_j q+b_j
\,\bigm|\,
q\in
[q^{\min}_{h,\pi^{-1}(j)},
q^{\max}_{h,\pi^{-1}(j)}]
\bigr\}
\\[-0.25em]
&\qquad\subseteq
[q^{\min}_{r,j},q^{\max}_{r,j}],
\qquad \forall j.
\end{aligned}
\label{eq:range}
\end{equation}
Here, $\alpha_j$, $\varepsilon_{p,j}$, and
$\varepsilon_{\ell,j}$ are the allowed axis-direction, axis-location,
and link-dimension tolerances, respectively. For each corresponding
joint, $s_j\in\{-1,+1\}$ specifies the selected positive
joint-coordinate direction, and $b_j$ is the fixed zero-position
offset determined during calibration. The affine mapping for joint $j$ is
\begin{equation}
q_{r,j}
=
s_j q_{h,\pi^{-1}(j)}
+
b_j.
\label{eq:componentwise_map}
\end{equation}
Equivalently, teleoperation uses the direct joint-space command mapping
\begin{equation}
\begin{aligned}
\mathbf{q}_r
&=
\mathbf{S}\mathbf{P}_{\pi}\mathbf{q}_h+\mathbf{b},
\\
\dot{\mathbf{q}}_r
&=
\mathbf{S}\mathbf{P}_{\pi}\dot{\mathbf{q}}_h,
\end{aligned}
\label{eq:joint_space_map}
\end{equation}
where $\mathbf{S}=\operatorname{diag}(s_1,\ldots,s_n)$ and
$\mathbf{b}=[b_1,\ldots,b_n]^T$ represent the fixed
coordinate-direction and zero-position calibration.
The absolute inner product in the axis-direction condition treats a
physical joint axis as an unoriented line. The sign $s_j$ separately
accounts for the selected positive joint-coordinate direction, while
$b_j$ accounts for the calibrated zero-position offset.

This mapping is affine in joint coordinates and does not require solving
task-space inverse kinematics during exoskeleton-to-robot command
transfer. Kinematically mismatched interfaces may instead require
task-space inverse kinematics or workspace projection, which can
introduce additional computation, sensitivity near singular
configurations, or command clipping for out-of-workspace targets. The
mechanically limited exoskeleton joint ranges were designed so that their
calibrated affine images remain within the corresponding robotic-hand
joint limits.

\subsection{Human-First MILE Exoskeleton}

Based on these design criteria, we developed an anthropometrically
informed, four-finger, 17-DoF articulated hand exoskeleton with
per-finger serial chains for the thumb, index, middle, and ring fingers.
The little finger is intentionally omitted as a system-level
simplification to reduce wearable width, mass, actuator and encoder
count, wiring, packaging complexity, and inter-finger interference risk.
This layout supports the evaluated tasks but may limit tasks that
explicitly require little-finger contact or independent little-finger
control.

For the three implemented non-thumb fingers, the distal and proximal
interphalangeal joints (DIP and PIP) are implemented as single-axis
flexion--extension hinges, while each metacarpophalangeal (MCP) joint
provides two orthogonal axes for flexion--extension and
adduction--abduction. The thumb provides five degrees of freedom for
opposition, comprising a single-axis interphalangeal (IP) joint and
two-axis carpometacarpal (CMC) and MCP joints. The physical exoskeleton
layout is shown in Fig.~\ref{fig:exploded_view}(a).

The exoskeleton dimensions are informed by the representative adult hand
dimensions shown in Fig.~\ref{fig:Isomorphic_comparison}(a). To reuse common
joint, encoder, and assembly modules, the three implemented non-thumb fingers
use standardized exoskeleton link dimensions consistent with the selected
joint-space correspondence. The thumb link dimensions and base offset are
extended as an engineering compromise to increase opposition clearance,
reduce mechanical interference, and enlarge the usable workspace for in-hand
manipulation. This choice is not presented as an anatomically optimal design.
Fingertip, thumb--index web-space, and wrist straps provide limited adjustment
for different hand and wrist sizes, but the device remains an ergonomic
approximation. Additional dimensional and fitting details are provided in
Supplementary Section~III.

\begin{figure*}[t]
	\centering
	\includegraphics{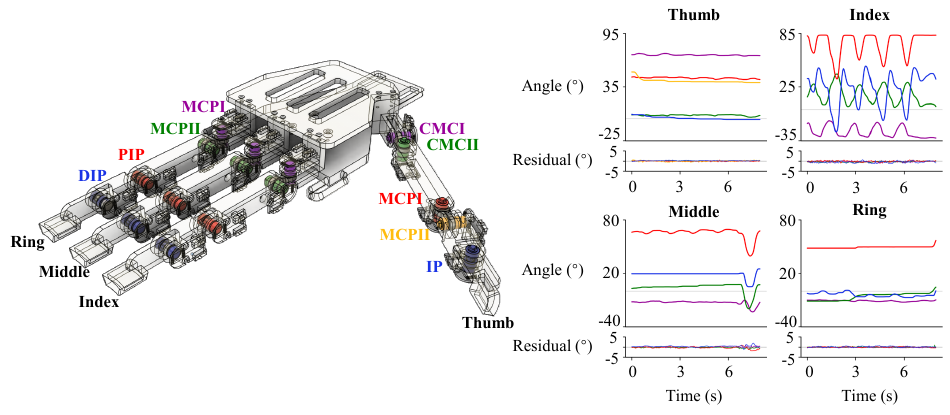}

\caption{Whole-exoskeleton multi-joint sensing validation against
mocap-derived references. For each finger, the upper panel shows
the encoder-measured joint-angle trajectories, while the lower strip
shows their residuals relative to the mocap-derived references. The
colors of the encoder lines match those of the corresponding joints in
the exoskeleton rendering.}

    \label{fig:joint_angle_comparision}
\end{figure*}

Two modular joint architectures realize these kinematics: rotational
hinges for single-axis joints and saddle hinges for two-axis joints.
Each saddle hinge separates orthogonal axes, producing the lateral
offsets in Fig.~\ref{fig:Isomorphic_comparison}(b) and (c).
These standardized hinge modules realize the selected exoskeleton
joint axes while reducing joint-specific packaging and assembly complexity.

As shown in Fig.~\ref{fig:exploded_view}(b), a common modular encoder
architecture comprising a radially magnetized rotor and a tunnel
magnetoresistance (TMR) sensor measures each of the 17 joint coordinates
around the corresponding joint shaft. The encoder module separates
the joint shaft/rotor, magnet, and TMR sensing unit, allowing the same
sensing architecture to be reused across the exoskeleton joints. Together, the modular joint-angle sensing architecture and the
calibrated affine exoskeleton-to-robot joint-space correspondence in
Section~III-A allow the measured exoskeleton coordinates to be
transferred to the MILE-Tac hand without solving task-space inverse
kinematics or nonlinear optimization online.

\subsection{MILE-Tac Hand with Fingertip Visuotactile Sensing}

As shown in Fig.~\ref{fig:exploded_view}(c), the MILE-Tac
hand is a four-fingered, 17-DoF robotic hand designed to share the
selected kinematic-tree topology and corresponding joint-axis arrangement
with the MILE exoskeleton. Its modular actuator packaging,
3D-printed fabrication, and maintenance draw on the design principles
of the LEAP Hand~\cite{shawLEAPHandLowCost2023}, while the
four-finger kinematic topology and fingertip visuotactile integration
are customized for MILE. The selected four-finger implementation limits
actuator count and the associated packaging, wiring, and
power-distribution complexity. Dynamixel XC330 actuators are arranged
to realize the corresponding joint axes, and the load-bearing
robotic-hand structures are fabricated from HP 3D High Reusability
PA 12.

To provide robot-side fingertip contact observations, each fingertip of the
MILE-Tac hand integrates a compact, custom-designed and fabricated
MILE fingertip visuotactile sensor module. As shown in
Fig.~\ref{fig:exploded_view}(d), each fingertip visuotactile sensor module comprises a deformable gel
layer, an acrylic plate, a side-mounted LED illumination module with its
driver, a camera module, and 3D-printed structural parts. The transparent
sensing layer uses Solaris\texttrademark{} silicone rubber (Smooth-On, Inc.) with
an empirically selected A:B mass ratio of 1:2. The ratio was selected
from preliminary formulation screening to balance compliance, optical
clarity, recovery, deformation range, and fabrication stability. Detailed
screening procedures and cyclic-loading observations are provided in
Supplementary Section~IV-B. The fine structural parts of the fingertip visuotactile sensor are printed in PLA, distinct from the PA 12 used for the load-bearing robotic-hand
structures. The modular layout separates the gel, illumination, imaging,
and structural components for assembly and maintenance, while the
compact optical packaging preserves camera standoff within the
fingertip envelope and adjacent-finger clearance. Cycling stability of
the assembled fingertip visuotactile sensor is evaluated in Section~\ref{sec:cyclic_endurance}.

\begin{table}[t]
\centering
\caption{MILE exoskeleton joint-angle sensing precision. MI denotes magnetic interference.}
\label{tab:precision_summary}
\renewcommand{\arraystretch}{1.15}
\begin{tabular}{l l c c}
\toprule
\multicolumn{2}{c}{\textbf{Setting}} & \textbf{MAE ($^\circ$)} & \textbf{MaxAE ($^\circ$)} \\
\midrule
\multirow{2}{*}{Single joint}
& No MI   & 0.33 & 0.81 \\
& With MI & 0.37 & 1.58 \\
\midrule
Multi-joint
& MILE exoskeleton  & \textbf{0.41} & \textbf{1.96} \\
\bottomrule
\end{tabular}
\end{table}

\section{System Performance Evaluation}
\subsection{Precision Evaluation}
\label{sec:precision_evaluation}
We evaluate the agreement between MILE exoskeleton joint-angle measurements
and mocap-derived joint-angle references at both the single-joint and assembled
multi-joint levels, with the evaluated settings and comparison metrics summarized
in Table~\ref{tab:precision_summary}.

\subsubsection{Setup}
The precision evaluation uses a 12-camera FZMotion Swift120A mocap
system. Rigid marker constellations are attached to the corresponding rigid exoskeleton links, and the resulting mocap-derived link orientations are used to construct joint-angle references characterizing exoskeleton-link motion.

\subsubsection{Metrics}
For joint $j$, let $\theta^{\mathrm{enc}}_j[k]$ and $\theta^{\mathrm{mocap}}_j[k]$ denote the encoder measurement and mocap-derived joint-angle reference, respectively, at retained synchronized sample $k$. We report
\begin{equation}
\mathrm{MAE}_j =
\frac{1}{N_j}
\sum_{k=1}^{N_j}
\left|
\theta^{\mathrm{enc}}_j[k]
-
\theta^{\mathrm{mocap}}_j[k]
\right|.
\end{equation}
\begin{equation}
\mathrm{MaxAE}_j =
\max_{1\leq k\leq N_j}
\left|
\theta^{\mathrm{enc}}_j[k]
-
\theta^{\mathrm{mocap}}_j[k]
\right|.
\end{equation}

Here, $k$ is the retained synchronized-sample index and $N_j$ is the number of retained samples for joint $j$. The reported multi-joint MAE is the arithmetic mean of $\mathrm{MAE}_j$ over the 17 evaluated joints, with each joint-wise metric computed on its retained synchronized interval; the reported multi-joint MaxAE is $\max_j \mathrm{MaxAE}_j$.

\subsubsection{Protocols}
(\romannumeral 1) \emph{Single-joint encoder precision with and without an
externally introduced magnetic perturbation:} A Dynamixel XC330 drives one
joint through cyclic flexion--extension. In the With-MI condition, a moving
external permanent magnet introduces a magnetic perturbation near the encoder;
the No-MI condition uses the same setup without this external magnet. Encoder
measurements are compared with mocap-derived joint-angle references under both
conditions. During teleoperation, the robotic-hand motors and exoskeleton
encoders are physically separated; together with the perturbation experiment,
this suggests that motor-induced magnetic interference is unlikely to dominate
the sensing error under the tested configuration.

(\romannumeral 2) \emph{Multi-joint full-exoskeleton evaluation:}
All 17 exoskeleton joint coordinates are recorded simultaneously during
dynamic free-finger motion of the assembled exoskeleton. The encoder
measurements are compared with mocap-derived joint-angle references
constructed from rigid marker constellations attached to the corresponding
exoskeleton links.

\subsubsection{Results}
Table~\ref{tab:precision_summary} summarizes the joint-angle sensing validation
of the MILE exoskeleton. The single-joint tests characterize encoder--mocap
agreement with and without the externally introduced magnetic perturbation; the
encoder operating principle and complete single-joint traces are provided in
Supplementary Figs.~S1 and S2 and Supplementary Video~2. The assembled MILE
exoskeleton shows a multi-joint MAE of $0.41^\circ$ relative to the
mocap-derived references, with a maximum absolute residual of $1.96^\circ$,
as shown in Fig.~\ref{fig:joint_angle_comparision}. Complete per-joint encoder
and mocap-derived reference trajectories are provided in Supplementary
Fig.~S3. These results characterize the exoskeleton joint-angle measurements
used for direct joint-space command transfer.

\subsubsection{Cyclic Endurance Evaluation}
\label{sec:cyclic_endurance}

To assess cycling stability under repeated use, we tested an encoder-equipped exoskeleton joint for 100,000 periodic motion cycles and analyzed a fingertip visuotactile cyclic-compression experiment comprising 2,300 cycles spanning approximately 11~h. The first 300 cycles were used for model training, and the remaining 2,000 cycles were used for endurance evaluation and visualization. Under the tested protocols, the joint-angle residual showed no increasing trend, and the measured force range of the visuotactile sensor changed only slightly. Detailed endurance protocols and results are provided in
Supplementary Section~IV-B.

\subsection{Teleoperation Performance Evaluation}
\label{sec:Teleoperation_Performance_Evaluation}

\begin{figure}[!t]
    \centering
    \includegraphics{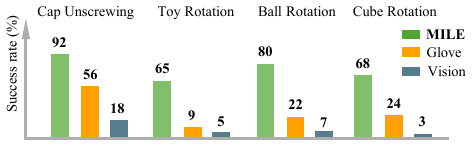}
    \caption{Task-level teleoperation success-rate comparison among the MILE
exoskeleton, the MANUS Quantum MetaGloves-based interface, and the
RealSense D435-based vision interface.}
    \label{fig:teleoperation result}
\end{figure}

We evaluate teleoperation on four contact-rich in-hand manipulation tasks: cap unscrewing, toy rotation, ball rotation, and cube rotation. Fig.~\ref{fig:teleoperation result} compares the task-level
success rates of the three interfaces. For each task, the same operator used all three interfaces.
For each interface--task condition, 100 formal trials were conducted
under a common 60-s time limit. All three interfaces commanded the same
MILE-Tac hand. Detailed task definitions and success criteria are provided
in Supplementary Section~V-A.

Additional teleoperated demonstrations, including varied grasping
tasks and continuous in-hand object reorientation, are provided in the
Supplementary Material and accompanying videos.

\subsubsection{Baselines}
We compare MILE with two retargeting-based interfaces: a MANUS Quantum
MetaGloves interface and an Intel RealSense D435 vision interface. The
MANUS interface uses glove-derived finger keypoints with
inverse-kinematics-based retargeting~\cite{shaw2024bimanual}, whereas
the vision interface uses RGB-based hand landmarks with
optimization-based kinematic retargeting~\cite{qin2023anyteleop}.
All three interfaces command the same MILE-Tac hand.
\rev{Supplementary Sections~VI-A and VI-B describe the baseline
implementations and pipeline timing, respectively.}

\subsubsection{Results}
Averaged over the four tasks, the MILE exoskeleton interface achieved a mean success rate of 76\%, compared with 28\% for the MANUS Quantum MetaGloves interface and 8\% for the RealSense interface, corresponding to absolute increases of 48 and 68 percentage points, respectively, as shown in Fig.~\ref{fig:teleoperation result}.

\begin{figure*}[t]
\centering
\includegraphics{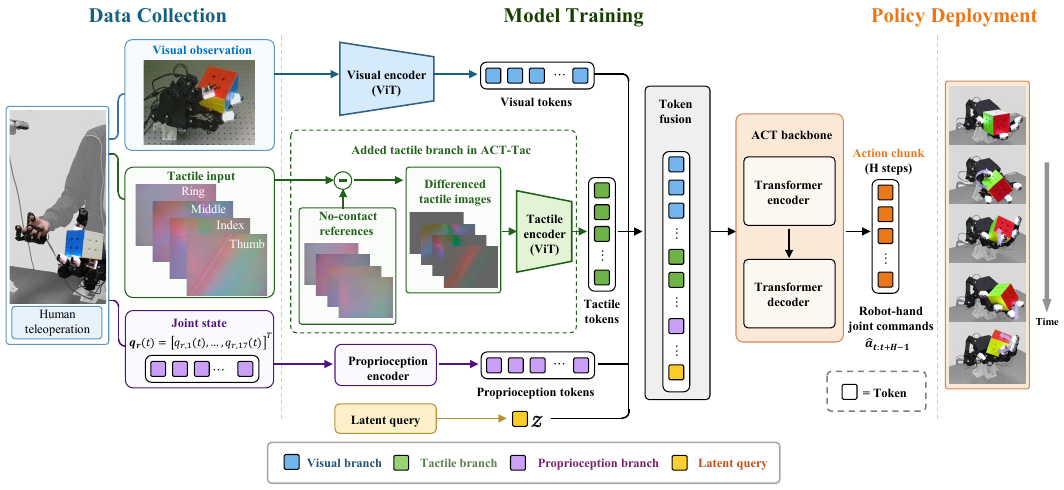}
\caption{ACT-Tac policy pipeline. Visual, tactile, proprioceptive, and latent-query tokens are fused by the ACT backbone to predict robot-hand joint-command chunks for policy deployment.}
\label{fig:pipeline}
\end{figure*}

\section{Application to Imitation Learning}
We evaluate whether the synchronized multimodal demonstrations collected by
MILE can support downstream learning across two representative but structurally
distinct policy backbones, ACT and DP, spanning transformer-based action chunking
and diffusion-based generative policy learning, respectively. The paired tactile
and non-tactile variants are used to assess whether the fingertip visuotactile
observations collected by MILE provide complementary, policy-relevant information
across these downstream learning frameworks.

\subsection{Demonstration Collection}
\label{sec:data_collection}

The in-hand demonstrations used for policy learning were collected through
the MILE exoskeleton--MILE-Tac teleoperation pipeline evaluated in
Section~\ref{sec:Teleoperation_Performance_Evaluation}. To extend MILE beyond hand-only dexterous teleoperation, we used a VIVE
Tracker to measure the operator's hand pose and commanded the Flexiv Rizon 4
arm to follow the corresponding end-effector pose, while the MILE exoskeleton
provided joint-space commands to the MILE-Tac hand.
The same arm--hand teleoperation setup was used to collect
demonstrations for both the fixed-bottle cap-unscrewing and sequential
potato-chip pick-and-place tasks.
Additional implementation details for this robotic-arm-mounted setup are
provided in Supplementary Section~VII-A. Because the
MANUS Quantum MetaGloves and RealSense interfaces were not evaluated with the MILE-Tac hand
mounted on a robotic arm under the same protocol, these experiments are not reported as a direct
teleoperation baseline comparison. Instead, they are used for training and evaluating autonomous
policies with and without fingertip tactile input, as reported in Table~\ref{tab:il_combined}.

\subsection{Experimental Setup and Evaluation Protocols}
\subsubsection{Hardware}

All streams are logged on a workstation running Ubuntu 22.04 with an
Intel Core i9\mbox{-}14900KF CPU and an NVIDIA RTX~4090 GPU.
A fixed RealSense D435 provides an external view of the hand--object
workspace. The sequential potato-chip task additionally uses a RealSense
D405 rigidly mounted on the MILE-Tac hand to provide an egocentric view
of the fingertip manipulation region. Detailed camera placement and the
corresponding experimental configurations are provided in Supplementary
Section~VII-B and Fig.~S6.

\subsubsection{Evaluation Tasks and Protocols}

Table~\ref{tab:il_combined} reports autonomous-policy performance on four in-hand manipulation tasks---cap unscrewing, ball rotation, toy rotation, and cube rotation---and two additional tasks using the MILE-Tac hand mounted on a robotic arm: fixed-bottle cap unscrewing and sequential potato-chip pick-and-place. The demonstrations for these two additional tasks were collected using the VIVE-Tracker-based teleoperation setup described in Section~\ref{sec:data_collection}. The four policy variants are ACT, ACT-Tac, DP, and DP-Tac, where the -Tac variants include fingertip tactile input. For each of the four in-hand tasks, each policy variant was evaluated in 30 trials using the same task definitions and 60-s time limits as the teleoperation evaluation in Section~IV-B. Each policy variant was also evaluated in 30 fixed-bottle cap-unscrewing trials using the MILE-Tac hand mounted on a robotic arm under a 30-s time limit. Detailed task definitions and success criteria for all evaluations are provided in Supplementary Section~V. For sequential potato-chip pick-and-place, each policy variant was
evaluated in ten continuously recorded sequences, each containing nine
individual-chip trials. Each chip constituted an independent policy
trial, yielding 90 trials per policy variant. Each trial used a 30-s
time limit, and success was counted per chip.

\subsection{Imitation Learning with Fingertip Tactile Input}

We use in-hand bottle-cap unscrewing as a representative task to describe
the policy inputs because it requires stable multi-contact and
contact-sensitive rotational manipulation.

\subsubsection{Sensing and Dataset}

For this task, 196 demonstrations were used for policy training
(Supplementary Video~1, Part III), and the corresponding data modalities
are summarized in Supplementary Fig.~S7. The dataset covers four manipulation strategies: thumb-only, index-only, middle-only, and thumb--index cooperative manipulation.
Each demonstration contains policy observations and the corresponding robot-hand joint-command targets. During teleoperation, the measured MILE-exoskeleton coordinates are mapped online through the calibrated affine joint-space correspondence in Eq.~(\ref{eq:joint_space_map}) to generate the commands sent to the MILE-Tac hand. These mapped robot-hand commands, rather than the raw exoskeleton-coordinate readings, are recorded as the demonstrated action targets. The policy observation at time $t$ comprises:
\begin{itemize}
\item \textbf{Vision} $\mathbf{I}_t^{\mathrm{vis}}$: a task-specific RGB or RGB-D observation of the hand--object workspace, depending on the task configuration.
\item \textbf{Tactile} $\mathbf{T}_t^{(f)}$: episode-referenced fingertip visuotactile images from fingertips $f\in\{1,2,3,4\}$.
\item \textbf{Proprioception} $\mathbf{q}_r(t)$: the measured joint positions of the MILE-Tac hand.
\end{itemize}
Let $\mathbf{a}_t$ denote the recorded 17-DoF robot-hand joint command used as the demonstrated action target.

At time \(t\), the
observation is
\begin{equation}
\mathbf{o}_t
=
\left(
\mathbf{q}_r(t),
\phi_{\mathrm{vis}}\!\left(\mathbf{I}_t^{\mathrm{vis}}\right),
\phi_{\mathrm{tac}}\!\left(\{\mathbf{T}_t^{(f)}\}_{f=1}^{4}\right)
\right).
\end{equation}
For the non-tactile variants, the tactile-encoding term is omitted.

Here, $\phi_{\mathrm{vis}}$ and $\phi_{\mathrm{tac}}$ denote the learned
visual and tactile encoders. The policy maps the resulting observation
to a robot-hand joint command,
\begin{equation}
\hat{\mathbf{a}}_t
=
\pi_{\theta}(\mathbf{o}_t),
\end{equation}
where $\hat{\mathbf{a}}_t$ is supervised by the recorded command
$\mathbf{a}_t$. During deployment, the predicted command is used as the
MILE-Tac hand joint target.

We instantiate Action Chunking Transformer (ACT) and Diffusion Policy (DP)
as policy backbones. As shown in Fig.~\ref{fig:pipeline}, ACT-Tac augments
the ACT policy with a tactile branch that encodes the four fingertip
visuotactile streams into tactile tokens and fuses them with
visual-observation tokens, robot-hand proprioceptive tokens, and the latent
query in the ACT backbone. \rev{The ACT Transformer predicts an action chunk from these tokens.
DP-Tac uses fused visual, tactile, and proprioceptive features
to condition a one-dimensional U-Net that denoises action sequences.
Both backbones output chunks of future robot-hand joint commands
in the same representation as the recorded commands.} To evaluate tactile input, we
compare \textbf{VP} (external scene-view observations + proprioception) and
\textbf{VPT} (VP + four fingertip visuotactile streams: internal-camera
images of contact and sensing-surface deformation).

\rev{Supplementary Section~VII-D details tactile preprocessing, data
synchronization, operator tactile visualization, feature fusion, and
policy training.}

\subsubsection{Results}

\begin{figure*}[t]
	\centering
	\includegraphics{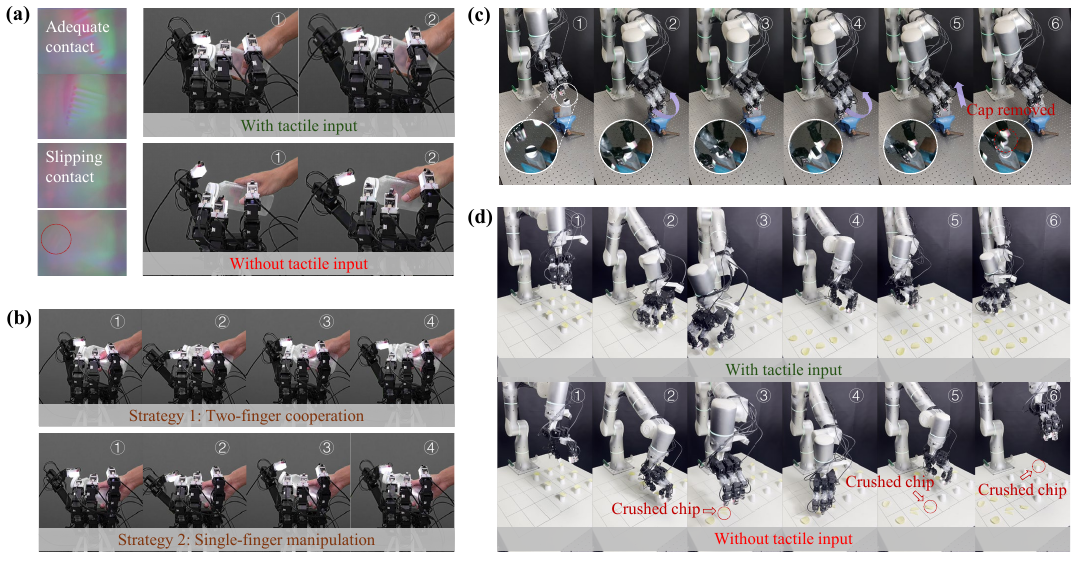}
\caption{Representative ACT-based autonomous manipulation results.
(a) Cap-unscrewing contact behavior with and without tactile input; the
red circle marks a weak tactile cue in the slipping-contact example.
(b) Two cap-manipulation behaviors observed during autonomous ACT-Tac unscrewing: two-finger cooperation and single-finger manipulation (Supplementary Video~3).
(c) Autonomous fixed-bottle cap unscrewing with the MILE-Tac hand
mounted on a robotic arm; the circular insets show the RGB observations
used by the policy (Supplementary Video~4).
(d) Sequential potato-chip pick-and-place with and without tactile
input; the red annotations mark crushed chips in the no-tactile trials
(Supplementary Video~4).
Complete ACT, ACT-Tac, DP, and DP-Tac success rates are reported in
Table~\ref{tab:il_combined}.}
	\label{fig:autonomous_extension}
\end{figure*}

Representative qualitative examples are shown in
Fig.~\ref{fig:autonomous_extension}, and complete quantitative results
are reported in Table~\ref{tab:il_combined}. Across the 12 evaluated
task--backbone pairs, the tactile-input variants achieved numerically
higher success rates than their corresponding non-tactile variants,
with an unweighted macro-average absolute success-rate difference of
14.7 percentage points. The improvements across the evaluated tasks are
consistent with fingertip tactile observations providing complementary
contact-state information for contact-rich manipulation.
Representative failure cases further illustrate the role of contact
information. Fig.~\ref{fig:autonomous_extension}(a) shows slipping contact
during cap unscrewing, while Fig.~\ref{fig:autonomous_extension}(d) shows
chip damage in no-tactile sequential potato-chip trials. Supplementary
Video~4 shows nine consecutively successful individual-chip ACT-Tac trials
within one continuous sequence across chips with varied shapes and
orientations. Together with the quantitative results, these examples
indicate that the fingertip visuotactile observations collected by MILE
provide complementary, policy-relevant contact information that can be
utilized by both ACT- and DP-based policies.

\section{Conclusion}

This work presents MILE as a human-first, constraint-driven approach to collecting multimodal demonstrations for dexterous manipulation. By establishing a calibrated joint-space correspondence between the wearable exoskeleton and the MILE-Tac hand, the system transfers measured operator joint coordinates directly to robotic-hand commands without online task-space IK-based retargeting. The teleoperation benchmark provides a system-level comparison under the implemented baseline configurations and supports the practical effectiveness of this hardware-level correspondence across the tested dexterous-hand tasks. These results position MILE as a human-first hardware-correspondence approach, complementing online retargeting-based interfaces and target-robot-driven hardware adaptation.

The downstream policy experiments provide complementary evidence:
across both ACT and DP backbones, policies using the MILE-collected
fingertip visuotactile observations achieved numerically higher success
rates than their paired non-tactile variants across all evaluated
task--backbone pairs. Together with the qualitative examples of slipping
contact during cap unscrewing and chip damage during sequential
potato-chip pick-and-place, these results indicate that the recorded
fingertip channel contains complementary contact information that can be
exploited by different policy architectures. Overall, MILE combines
human-side anatomical and wearability considerations with direct
joint-space correspondence for dexterous-hand teleoperation and data
collection.

\begin{table}[t]
\centering
\caption{Autonomous-policy task success with and without fingertip
tactile input.}
\label{tab:il_combined}
\renewcommand{\arraystretch}{1.12}
\setlength{\tabcolsep}{1.5pt}
\footnotesize
\begin{tabular}{@{}p{0.56\columnwidth}cccc@{}}
\toprule
\textbf{Task} &
\textbf{ACT} &
\mbox{\textbf{ACT-Tac}} &
\textbf{DP} &
\mbox{\textbf{DP-Tac}} \\
\midrule
\emph{MILE-Tac hand only} & & & & \\
In-hand Cap Unscrewing
  & 23/30
  & 27/30
  & 9/30
  & 11/30 \\
Ball Rotation
  & 18/30
  & 25/30
  & 7/30
  & 10/30 \\
Toy Rotation
  & 13/30
  & 17/30
  & 5/30
  & 6/30 \\
Cube Rotation
  & 7/30
  & 9/30
  & 5/30
  & 6/30 \\
\midrule
\emph{MILE-Tac hand mounted on a robotic arm} & & & & \\
Fixed-Bottle Cap Unscrewing
  & 26/30
  & 29/30
  & 23/30
  & 28/30 \\
Sequential Potato-Chip Pick-and-Place
  & 51/90
  & 79/90
  & 47/90
  & 82/90 \\
\bottomrule
\end{tabular}
\end{table}

The current MILE prototype has limitations in hardware and autonomous performance. Its four-finger, 17-DoF configuration omits the little finger, and exoskeleton fit depends on hand size, strap placement, and attachment compliance. The fingertip visuotactile modules add packaging and bandwidth demands, while the system provides no active force or tactile feedback to the operator. Autonomous policy success also leaves room for improvement across the evaluated tasks. Future work will investigate lighter and more adaptable wearable hardware with broader finger coverage, more compact and integrated visuotactile sensors with bandwidth-efficient data transmission for multifingered hands, improved visual--tactile representations and policy fusion, and active operator-side force or tactile feedback.

\section{Acknowledgment}
The authors thank Yunfan Zhang, Qianyou Zhao, Yongyao Li, Xu Song, and Zheng Wang for advice on hardware design and learning policies; and Longyan Wu, Yueshi Dong, Jiapeng He, Nianzu Lv, Yuxuan Wu, Yutong Pei, Jinnuo Zhang, Zhenle Liu, Boyang Peng, and Junjie Xia for assistance with data collection and hardware renderings.

% Combined arXiv version: main manuscript, supplementary material, shared references.
\input{supplementary}

\clearpage
\IEEEtriggeratref{21}
\bibliographystyle{IEEEtran}
\bibliography{IEEEabrv,TMECH2025}

\end{document}

%% file: supplementary.tex
% This file is included by root.tex after the complete main manuscript.
% It intentionally contains no document class, package imports, bibliography,
% or document environment of its own.
\clearpage
\setcounter{section}{0}
\setcounter{subsection}{0}
\setcounter{subsubsection}{0}
\setcounter{figure}{0}
\setcounter{table}{0}
\renewcommand{\thefigure}{S\arabic{figure}}
\renewcommand{\thetable}{S\Roman{table}}
% The visible counters restart for the supplement, while the hyperref anchors
% stay unique within the combined PDF.
\renewcommand*{\theHsection}{supplementary.\arabic{section}}
\renewcommand*{\theHfigure}{supplementary.\arabic{figure}}
\renewcommand*{\theHtable}{supplementary.\arabic{table}}
\twocolumn[
\begin{@twocolumnfalse}
\begin{center}
{\LARGE\bfseries Supplementary Material for ``MILE: A Mechanically
Isomorphic Hand Exoskeleton and Visuotactile Robotic Hand for Data Collection
in Dexterous Manipulation''\par}
\vspace{1em}
{\normalsize Jinda Du, Jieji Ren, Qiaojun Yu, Ningbin Zhang, Yu Deng,
Xingyu Wei, Yufei Liu, Guoying Gu, and Xiangyang Zhu\par}
\vspace{1.5em}
\end{center}
\end{@twocolumnfalse}
]

\section{Supplementary Videos}

\textbf{Supplementary Video 1: MILE System Operation and Data Collection.}
This video presents three parts: exoskeleton donning, human teleoperation,
and manipulation demonstrations. Part I shows the donning procedure of the
MILE exoskeleton.
Part II presents sequential teleoperation demonstrations for cap unscrewing
using different finger strategies.
The teleoperation footage is shown at real-time playback.
Part III presents external-view visual recordings from the 196 bottle-cap
demonstrations used for policy training and illustrates the diversity of the
collected manipulation trajectories.

\textbf{Supplementary Video 2: Precision Experiment.}
This video shows the encoder precision experiment with and
without an externally introduced magnetic perturbation. The on-screen labels
identify the motion capture (mocap) marker, encoder, external magnet, and motor.
During cyclic motion, a small handheld permanent magnet commonly used
in classroom physics demonstrations is slowly moved toward the encoder
to introduce a slowly varying external magnetic perturbation. Under the
tested motion and spacing, the encoder remains closely aligned with the
mocap-derived reference, with no obvious interference-induced degradation.

\textbf{Supplementary Video 3: Autonomous In-Hand Manipulation (ACT-Tac).}
This video combines autonomous ACT-Tac policy rollouts for cap unscrewing, cube rotation and
toy rotation shown at real-time playback. The cap-unscrewing segment shows representative autonomous cap-unscrewing
inference examples for thumb--index, thumb-only, index-only, and middle-only
manipulation strategies.
The cube-rotation segment shows cube rotation from an initial condition with
any non-red face facing upward to a red-face-up goal condition; the final
in-plane orientation is unconstrained.
The toy-rotation segment shows toy rotation from a front-side-up initial
condition to a back-side-up goal condition; the final in-plane orientation is
unconstrained. The visible white item on the toy is the manufacturer's product
tag and is not part of the task or evaluation.

\textbf{Supplementary Video 4: Autonomous Arm--Hand Manipulation (ACT-Tac).}
This video combines autonomous ACT-Tac policy rollouts for fixed-bottle
cap unscrewing and sequential potato-chip pick-and-place using the MILE-Tac
hand mounted on a robotic arm, with the manipulation footage shown at
real-time playback.
For fixed-bottle cap unscrewing, the bottle is fixed in the environment.
The sequential potato-chip segment shows a representative successful sequence
of nine consecutive individual-chip ACT-Tac inference trials. The manipulation
sequence is recorded continuously without cuts.

\section{Comparison of Kinematic-Correspondence Strategies}

Table~\ref{tab:supp_correspondence} compares representative hardware
strategies for reducing kinematic mismatch between an operator-side
interface and the target robot. The comparison focuses on the design basis,
the level at which kinematic correspondence is established, and the role of
human-side constraints.

GELLO~\cite{wu2024gello} starts from the target robot and uses a scaled
controller with corresponding joints and links, enabling direct mapping
between controller and robot joint coordinates. DexUMI~\cite{xu2025dexumi}
reduces kinematic mismatch through target-hand-driven hardware adaptation,
in which a wearable exoskeleton is optimized to match the target hand's
fingertip kinematics while allowing other aspects of the exoskeleton linkage
geometry to differ from those of the target hand where needed to satisfy
wearability constraints.

MILE follows a different design direction. Human-hand anatomy and
wearability requirements are prioritized in defining the wearable exoskeleton
and the selected 17-DoF topology and joint-axis arrangement, while robot-side
realizability constrains the implementation of the MILE-Tac hand. The
MILE-Tac hand is designed to share this selected topology and corresponding
joint-axis arrangement, with its dimensions and packaging adapted for
robot-side actuation and fingertip sensing. The resulting calibrated
joint-coordinate correspondence enables direct exoskeleton-to-robot
joint-space command transfer without online task-space IK-based retargeting.

\begin{table*}[!t]
\centering
\caption{Comparison of representative kinematic-correspondence strategies.}
\label{tab:supp_correspondence}
\small
\renewcommand{\arraystretch}{1.15}
\setlength{\tabcolsep}{4pt}
\begin{tabular}{@{}
p{0.09\textwidth}
p{0.17\textwidth}
p{0.41\textwidth}
p{0.28\textwidth}
@{}}
\hline
\textbf{System} &
\textbf{Design basis} &
\textbf{Kinematic correspondence} &
\textbf{Human-side constraints} \\
\hline

GELLO~\cite{wu2024gello} &
Target robot arm &
Scaled joint/link correspondence. &
No wearable-hand fitting. \\

DexUMI~\cite{xu2025dexumi} &
Target robotic hand &
Fingertip-kinematics matching. &
Wearability in optimization. \\

\textbf{MILE} &
\textbf{Wearable exoskeleton} &
\textbf{Shared selected topology and joint-axis arrangement.} &
\textbf{Human-side constraints prioritized.} \\

\hline
\end{tabular}
\end{table*}

\begin{figure*}[!t]
\centering
\begin{minipage}[t]{0.48\textwidth}
\centering
\includegraphics[width=\linewidth]{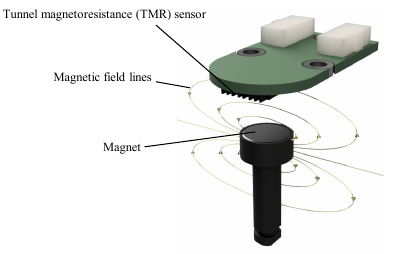}
\caption{Operating principle of the contactless tunnel magnetoresistance (TMR) joint-angle encoder. A radially magnetized target magnet is mounted coaxially near the TMR sensor, such that joint rotation changes the local magnetic-field direction measured by the sensor.}
\label{fig:supp_encoder_principle}
\end{minipage}
\hfill
\begin{minipage}[t]{0.48\textwidth}
\centering
\includegraphics[width=\linewidth]{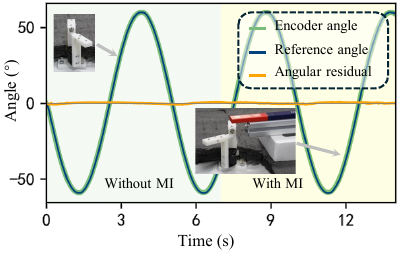}
\caption{Single-joint encoder precision test with and without an externally introduced magnetic perturbation. Here, MI denotes magnetic interference. The encoder angle follows the mocap-derived reference under both tested conditions, while the angular residual remains small under the tested setup.}
\label{fig:supp_single_encoder}
\end{minipage}
\end{figure*}

\section{Design Dimensions and Wearability Considerations}

The MILE exoskeleton adopts standardized index, middle, and ring
finger link dimensions so that common joint, encoder, and assembly
modules can be reused across the three implemented non-thumb fingers.
This choice supports hardware modularity and is consistent with the
selected exoskeleton-to-robot joint-space correspondence. The corresponding structural dimensions of the MILE exoskeleton and MILE-Tac hand annotated in Fig.~2(b) and (c) of the main manuscript are
summarized in Table~\ref{tab:supp_dimensions}. The MILE-Tac dimensions
are uniformly scaled from the corresponding exoskeleton dimensions by
the implemented geometric scale factor $\lambda=9/5$.

\begin{table}[!t]
\centering
\caption{Corresponding structural dimensions of the MILE exoskeleton and
MILE-Tac hand.}
\label{tab:supp_dimensions}
\small
\renewcommand{\arraystretch}{1.10}
\setlength{\tabcolsep}{10pt}
\begin{tabular}{lll}
\hline
$i$ & $L_i^E$ (mm) & $L_i^R$ (mm) \\
\hline
1 & 120 & 216 \\
2 & 25 & 45 \\
3 & 30 & 54 \\
4 & 24.72 & 44.5 \\
5 & 44 & 79.2 \\
6 & 11 & 19.8 \\
\hline
\end{tabular}
\end{table}

The wearable exoskeleton weighs approximately 200~g, excluding the
external VIVE Tracker used only when collecting demonstrations with the
MILE-Tac hand mounted on a robotic arm. The fingertip, thumb--index
web-space, and wrist straps provide limited adjustment for different hand
and wrist sizes. The straps couple the user's distal links to the
corresponding exoskeleton links and help stabilize the base of the
mechanism relative to the hand, reducing local lift-off during operation.
However, the attachment compliance, strap placement, and hand-size
variation remain practical factors in wearable use. Larger mismatch can
reduce usable range, joint-axis alignment, comfort, or attachment
stability, and no universal hand-to-exoskeleton fit threshold is claimed
in this work.

\section{Precision and Reliability Validation}

\subsection{TMR Encoder Operating Principle and Precision Validation}

\begin{figure*}[t]
\centering
\includegraphics[width=\textwidth]{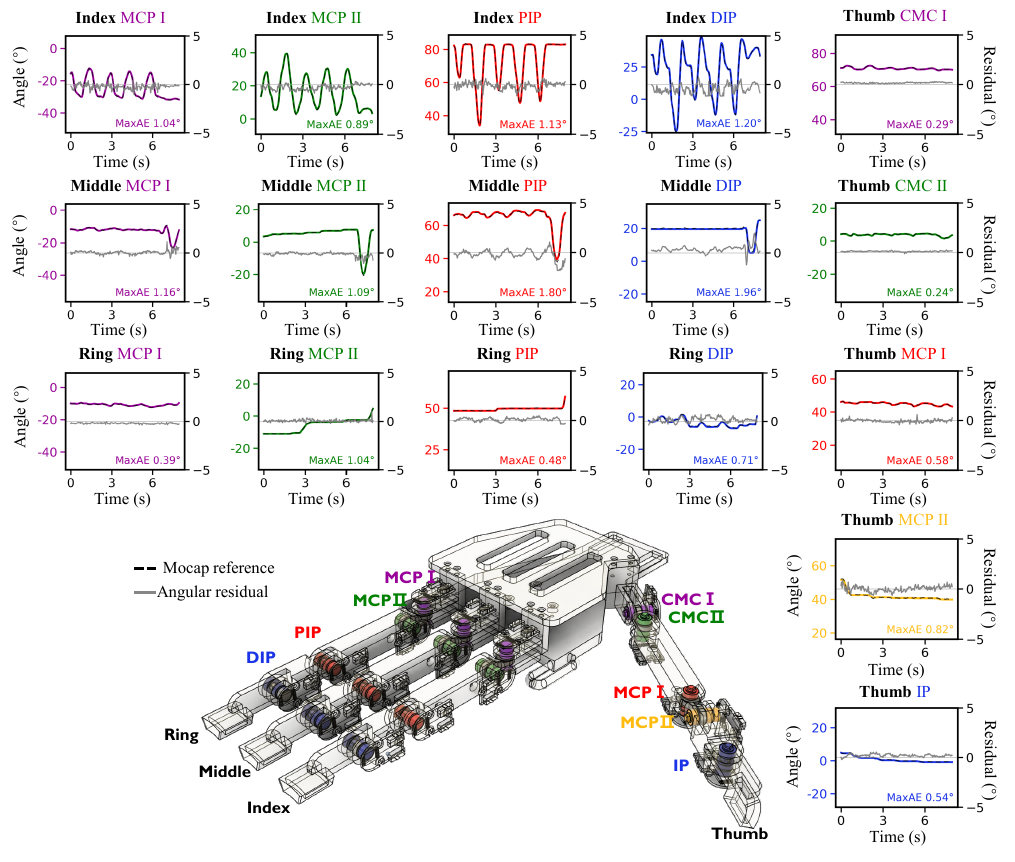}
\caption{Full-exoskeleton multi-joint encoder–mocap comparison. Encoder measurements are compared with
mocap-derived joint-angle references for all 17 joints during dynamic
free-finger motion. The colored solid lines show the encoder
measurements, the black dashed lines show the mocap-derived joint-angle
references, and the gray lines on the right axes show the angular
residuals. The colors of the encoder lines match those of the
corresponding joints in the exoskeleton rendering.
The per-panel annotations show the maximum absolute residual for each joint.}
\label{fig:supp_exoskeleton_precision}
\end{figure*}

\subsubsection{Operating Principle}

Fig.~\ref{fig:supp_encoder_principle} illustrates the operating principle of the joint-angle sensing module. A miniature radially magnetized permanent magnet, approximately 3~mm in diameter, is mounted coaxially at the end of the rotating joint shaft, with an approximately 1~mm air gap between the magnet and the tunnel magnetoresistance (TMR) sensor. As the shaft rotates, the orientation of the local magnetic field at the sensor changes, enabling contactless measurement of the joint angle.

Because the target magnet is positioned close to the sensing element, its field is expected to dominate the local field measured by the sensor. The field of the miniature magnet decays rapidly with distance. Therefore, cross-talk between adjacent modules is expected to remain limited under the implemented spacing. During teleoperation, the robotic-hand motors are physically separated from the operator-side exoskeleton encoders, further limiting exposure of the encoders to motor-generated magnetic fields. These considerations apply to the implemented configuration and do not imply immunity to arbitrary external magnetic interference.

\subsubsection{Single-Joint Precision Validation}

Fig.~\ref{fig:supp_single_encoder} presents the single-joint precision experiment with and without an externally introduced magnetic perturbation. A Dynamixel XC330 drove the encoder-equipped joint through cyclic flexion--extension, while the encoder measurement was compared with the corresponding mocap-derived joint-angle reference. In the ``With MI'' condition, a moving external permanent magnet introduced a magnetic perturbation near the encoder; the ``Without MI'' condition was conducted without this external perturbing magnet. Here, MI denotes magnetic interference.

Under both tested conditions, the encoder measurement remained closely aligned with the mocap-derived reference. The corresponding quantitative errors are summarized in Table~I of the main manuscript.

\subsubsection{Full-Exoskeleton Multi-Joint Precision Validation}

Fig.~\ref{fig:supp_exoskeleton_precision} presents the simultaneously
recorded encoder and mocap-derived reference trajectories for all 17
measured joints of the assembled MILE exoskeleton during dynamic
free-finger motion. The FZMotion Swift120A mocap system used in this experiment has a
specified angular error below $0.1^\circ$. The reported encoder--mocap MAE
therefore reflects the deviation between the encoder measurements and the
mocap-derived joint-angle references under this measurement setup. The overall multi-joint encoder--mocap MAE is $0.41^\circ$, with a
maximum absolute residual of $1.96^\circ$, as summarized in Table~I of
the main manuscript.

\begin{table*}[t]
    \centering
    \caption{Endurance statistics of the TMR encoder and visuotactile sensor.}
    \label{tab:supp_endurance}
    \footnotesize

    \textbf{(a) TMR encoder}\\[3pt]
    \setlength{\tabcolsep}{12pt}
    \begin{tabular}{@{}lccc@{}}
    \hline
    Metric & First 10 cycles & Final 10 cycles & Entire test \\
    \hline
    MAE
    & \(0.361^\circ\)
    & \(0.349^\circ\)
    & \(0.352^\circ\) \\
    Maximum absolute residual
    & \(1.930^\circ\)
    & \(1.074^\circ\)
    & \(1.930^\circ\) \\
    Mean residual
    & \(-0.005^\circ\)
    & \(0.032^\circ\)
    & -- \\
    \hline
    \end{tabular}

    \vspace{8pt}

    \textbf{(b) Visuotactile force prediction}\\[3pt]
    \setlength{\tabcolsep}{12pt}
    \begin{tabular}{@{}lccc@{}}
    \hline
    Metric & First 10 cycles & Final 10 cycles & Entire evaluation \\
    \hline
    MAE
    & 0.009~N
    & 0.059~N
    & 0.019~N \\
    RMSE
    & 0.013~N
    & 0.062~N
    & 0.028~N \\
    Mean prediction error
    & \(-0.001\)~N
    & \(-0.059\)~N
    & \(-0.013\)~N \\
    Maximum absolute error
    & 0.070~N
    & 0.133~N
    & 0.133~N \\
    \hline
    \end{tabular}
\end{table*}

\subsection{Cyclic Endurance Evaluation}
\label{sec:supp_endurance}

\subsubsection{TMR Encoder Endurance}

\textbf{Setup and protocol.}
A motorized testbed drove an encoder-equipped MILE joint module through 100,000 periodic motion cycles.
Each cycle lasted approximately 3~s, corresponding to a total test duration of approximately 83.3~h.
The acquisition rate was 50~Hz. The XC330 internal motor encoder was used as the reference angle source, and the TMR-based encoder integrated
into the MILE joint was evaluated against this reference. Before the endurance test, the magnet was fixed to the rotating shaft to prevent relative sliding during repeated cyclic motion.

\begin{figure*}[!t]
    \centering
    \includegraphics{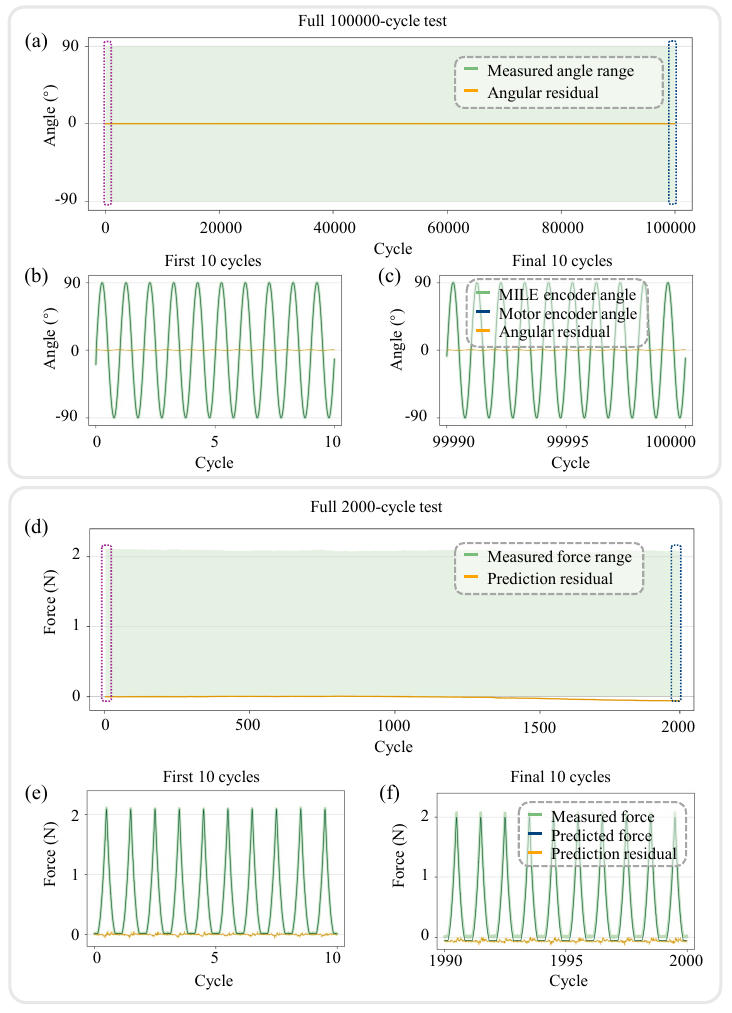}
    \caption{Endurance evaluation of the TMR encoder and visuotactile sensor. (a) Complete 100,000-cycle encoder test.
    The pale-green band denotes the per-cycle measured angle range, and the orange curve denotes the calibrated signed angular
    residual relative to the XC330 actuator-encoder reference. (b), (c) MILE-encoder and reference trajectories during the first
    and final ten cycles, respectively, under the same offset compensation. (d) 2,000-cycle visuotactile-sensor endurance-evaluation
    segment. The pale-green region denotes the measured force range within each ten-cycle bin, and the orange curve denotes the mean
    force-prediction residual in the same bin. (e), (f) Measured and predicted force trajectories during the first and final ten compression cycles
    of the endurance-evaluation segment, respectively. Detailed protocols and statistics are reported in
    Section~\ref{sec:supp_endurance} and Table~\ref{tab:supp_endurance}.}
    \label{fig:endurance}
\end{figure*}

\textbf{Data processing.}
A single initial angular offset between the MILE encoder and the XC330 actuator encoder was estimated from the first
100 cycles and then applied to the complete 100,000-cycle dataset. No additional re-zeroing or recalibration was performed
for the final cycles. The signed residual was calculated as the compensated MILE-encoder angle minus the XC330 actuator-encoder
reference and wrapped into the range $[-180^\circ,180^\circ)$ to avoid angular discontinuities. Because plotting
the sample-level angle trajectories from all 100,000 cycles in a single panel would obscure the long-term behavior,
Fig.~\ref{fig:endurance}(a) summarizes each cycle using the minimum and maximum measured angles over that cycle.
In Fig.~\ref{fig:endurance}(a), the pale-green band represents the per-cycle measured angle range over the complete
100,000-cycle test, while the orange curve shows the calibrated signed angular residual relative to the XC330
actuator-encoder reference. Fig.~\ref{fig:endurance}(b) and (c) show the MILE-encoder and actuator-encoder reference
trajectories during the first and final ten cycles, respectively, under the same offset compensation.

\textbf{Metrics and results.}
Table~\ref{tab:supp_endurance}(a) summarizes the encoder residual statistics. The first- and final-ten-cycle MAEs were similar,
changing from $0.361^\circ$ to $0.349^\circ$, and the complete-test MAE was $0.352^\circ$. Together with Fig.~\ref{fig:endurance}(a),
these results indicate no increasing residual trend over 100,000 cycles under the tested protocol. Because the same offset estimated
from the first 100 cycles was applied throughout the complete dataset, the small first-versus-final differences are consistent with
ordinary cycle-to-cycle variation under the fixed offset compensation.

\subsubsection{Visuotactile Sensor Endurance}

\textbf{Elastomer formulation.}
The sensing layer was fabricated from Solaris\texttrademark{} silicone rubber (Smooth-On, Inc.). Lower apparent compression
stiffness was considered desirable because it can produce greater deformation under comparable contact loading, thereby increasing
the magnitude of observable image changes. However, end-to-end tactile performance also depends on illumination, camera settings,
optical geometry, image processing, and calibration, while excessive compliance can reduce the usable force range and lead to early
saturation, bottoming-out, or residual deformation. In preliminary formulation screening, ten A:B mass ratios
(1:0.7, 1:1, 1:1.3, 1:1.5, 1:1.7, 1:2, 1:2.3, 1:2.5, 1:2.7, and 1:3) were each subjected to three compression loading cycles.
Force--displacement responses acquired under a common test configuration were used to compare relative compliance: A:B = 1:0.7 was
comparatively stiffer, and increasing the proportion of component B generally produced a more compliant response. Separate blends of
Solaris with other silicone materials became visibly turbid and were excluded as unsuitable for optical tactile imaging. These blends
were not counted as additional A:B ratios. Optical clarity and elastic recovery were assessed qualitatively, and the selection also
considered usable deformation range, obvious residual deformation, tackiness, handling characteristics, and fabrication stability.
Formulations with component-B proportions above approximately A:B = 1:2.5 exhibited tackiness and slower or incomplete recovery during
preliminary handling and repeated-loading observations. We therefore selected A:B = 1:2 as a practical low-stiffness compromise among
mechanical responsiveness, usable deformation range, optical clarity, elastic recovery, and fabrication stability. This process constituted
discrete engineering screening rather than continuous material optimization or standardized Young's-modulus characterization. The
endurance test reported below evaluates a fingertip visuotactile sensor fabricated using this selected formulation.

\textbf{Setup and protocol.}
The cyclic-compression experiment comprised 2,300 cycles spanning
approximately 11~h. The first 300 cycles were used for model training,
while the remaining 2,000 cycles were used for endurance evaluation and
visualization. The measured force was
used to evaluate the cyclic force response, and the tactile images were
used as input to a ResNet-based force-prediction model. The elastomer
layer was approximately 1.2~mm thick, and the nominal compression depth
was approximately 1~mm, corresponding to approximately 83\% of the
nominal layer thickness. The fingertip visuotactile sensor was compressed using a
cylindrical indenter with a diameter of approximately 7--8~mm on a
universal testing machine (68SC-2, Instron).

\textbf{Data processing.}
Because plotting all sample-level force trajectories in a single
panel would obscure the long-term behavior,
Fig.~\ref{fig:endurance}(d) summarizes the 2,000-cycle
endurance-evaluation segment using consecutive ten-cycle bins. The pale-green
region shows the measured force excursion within each bin, while the
orange curve shows the mean signed force-prediction residual over the
same bin. Fig.~\ref{fig:endurance}(e) and (f) provide magnified views of
representative cycles near the beginning and end of the
endurance-evaluation segment, respectively.

\textbf{Metrics and results.}
Table~\ref{tab:supp_endurance}(b) summarizes the force-prediction statistics over the 2,000-cycle endurance-evaluation segment. The measured cyclic force excursion remained broadly stable over the endurance-evaluation segment. The ResNet-based force predictor achieved an overall MAE of 0.019~N over the 2,000-cycle endurance evaluation. The MAE increased from 0.009~N during the first ten evaluation cycles to 0.059~N during the final ten evaluation cycles. However, the final error remained small compared with the approximately 2.1~N cyclic measured force range. The final residuals were dominated by a signed bias, with a final mean prediction error of $-0.059~\mathrm{N}$. Under the convention that prediction error is predicted force minus measured force, this corresponds to average underestimation. \rev{The prediction bias may reflect silicone viscoelasticity and
residual deformation, as cyclic-compression studies report stress
relaxation, hysteresis, and stress
softening~\cite{persson2022cycliccompression}.
Changes in material properties or illumination may also alter
tactile-image features and affect image-based force
prediction~\cite{chen2024deepdomainforce}.
The formulation screening above considers compliance and recovery,
and the modular gel layer permits replacement
(main manuscript, Section~III-C).
For downstream policy inputs, no-contact reference differencing
is intended to reduce static appearance offsets (Section~VII-D).
The observed bias suggests that calibration checks may be needed
during prolonged use.}

\begin{figure*}[!t]
    \centering
    \includegraphics{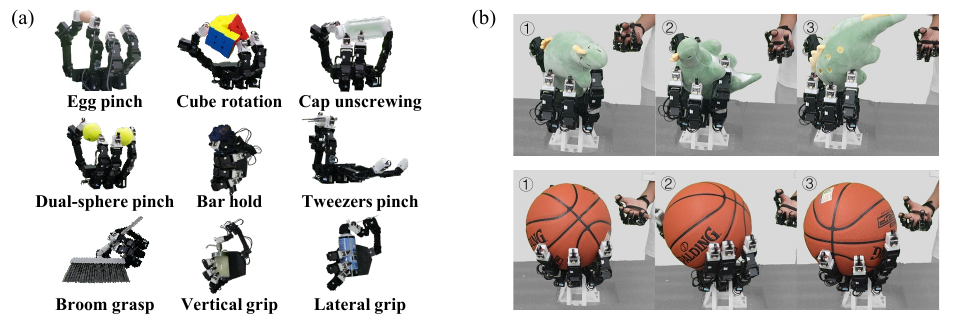}
    \caption{Additional MILE grasping and teleoperation demonstrations.
    (a) Representative grasp configurations and selected dexterous
    manipulation examples involving objects with different geometries and
    contact requirements. (b) Continuous teleoperation sequences for
    reorienting a complex-shaped toy and a basketball.}

    \label{fig:supp_teleoperation_demonstration}
\end{figure*}

\section{Manipulation Task Protocols}
\label{sec:supp_task_protocols}

\subsection{Hand-Only Tasks}

The following task definitions and success criteria were used for both the teleoperation benchmark in Section IV-B of the main manuscript and the autonomous-policy evaluation in Section V-B.2.

\textbf{Cap unscrewing.} A trial was considered successful if the bottle
cap was completely removed within 60~s. An assistant presented and
stabilized the bottle at varying poses within the robotic hand's
reachable workspace.

\textbf{Ball rotation.} A trial was considered successful if the ball completed at least one full revolution about the target axis within 60~s while remaining grasped and not dropped. The completed revolution was judged using a pencil reference line marked on the ball surface. Continued rotation after one full revolution did not invalidate the trial.

\textbf{Toy rotation.} The toy started from a front-side-up initial condition. A trial was considered successful if the toy reached a back-side-up condition within 60~s. The final in-plane orientation was unconstrained.

\textbf{Cube rotation.} The cube started with any non-red face upward. A trial was considered successful if the red face was upward within 60~s. The final in-plane orientation was unconstrained.

\subsection{Robotic-Arm-Mounted Tasks}

\textbf{Fixed-bottle cap unscrewing.} The bottle was fixed in the environment. Each policy variant was evaluated in 30 trials under a 30-s time limit, and a trial was considered successful if the cap was completely removed within 30~s.

\textbf{Sequential potato-chip pick-and-place.} Each individual chip manipulation was treated as one trial. Each policy variant was evaluated in ten continuously recorded sequences, each containing nine individual-chip trials. A new policy rollout was initiated for each chip and executed one complete pick-and-place attempt. Each individual-chip trial used a 30-s time limit. A trial was considered successful if the policy picked up one chip, transported it to the designated target location, and placed it within the target area within 30~s without dropping or crushing it. Failures included failed pickup, dropping, crushing or visible breakage, transport failure, placement substantially outside the target area, and exceeding the 30-s time limit. Failed trials remained included in the success-rate denominator.

\section{Teleoperation Performance Evaluation}

\subsection{Teleoperation Baseline Implementation}

The MANUS glove and vision-based baselines were adapted to the MILE-Tac hand
using its robot-specific kinematic model, active joints, target links, and
joint limits. \rev{For each baseline, the retargeting scale factor was set empirically
during preliminary teleoperation, and all settings were fixed before
the formal evaluation and used across all four benchmark tasks.}

\textit{MANUS glove baseline.}
The glove-based baseline used MANUS Quantum MetaGloves with MANUS Core
SDK 2.4~\cite{shaw2024bimanual}. Before teleoperation, the glove was calibrated to the operator's
hand using the manufacturer-provided calibration procedure in MANUS Core.
The RawSkeleton stream provided by MANUS Core was used for retargeting. For the thumb, index, middle, and ring fingers, the fingertip and an
adjacent distal keypoint were extracted as task-space targets. Because
MILE-Tac contains these four fingers, eight target points were used in
total, while the little-finger targets were omitted. The resulting targets were associated
with the corresponding links of the MILE-Tac hand and used as task-space
references for inverse kinematics. The MANUS target coordinates were uniformly scaled by a factor of 1.6.
The robot joint configuration was obtained using damped-least-squares
inverse kinematics in PyBullet, with a maximum of 50 iterations and a
residual threshold of $10^{-4}$.

\begin{figure*}[!t]
    \centering
    \includegraphics{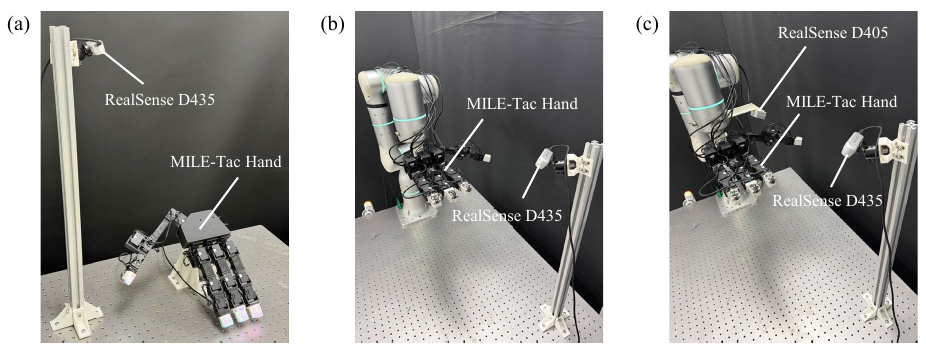}
    \caption{Camera configurations used in the manipulation
    experiments. (a) Hand-only manipulation tasks using a fixed external
    RealSense D435 oriented toward the hand--object workspace.
    (b) Robotic-arm-mounted fixed-bottle cap unscrewing using a fixed
    external RealSense D435 for a third-person workspace view.
    (c) Sequential potato-chip pick-and-place using the fixed external
    RealSense D435 together with an additional RealSense D405 mounted on
    the dorsal side of the MILE-Tac hand and oriented toward the fingertip
    manipulation region.}
    \label{fig:supp_camera_configuration}
\end{figure*}

\textit{Vision baseline.}
The vision-based baseline used an Intel RealSense D435 positioned approximately 0.5--1.0~m in front of the operator and facing the operating hand. During teleoperation, the operator's hand remained within the camera field of view, with the palm approximately facing the camera. MediaPipe Hands~\cite{48292} estimated hand landmarks, which were used as input to the retargeting pipeline. The MILE-Tac configuration used vector-based retargeting based on the formulation in~\cite{qin2023anyteleop}. For the thumb, index, middle, and ring fingers, the objective contained 12 local phalanx vectors,
corresponding to three consecutive bone vectors per finger, together with
all six pairwise fingertip vectors to preserve the relative multi-finger
configuration. The retargeting problem was solved using SLSQP. The formal
evaluation used a scaling factor of 1.5, a previous-frame regularization
coefficient of 0.001, a low-pass filtering coefficient of 0.2, and enabled
MILE-Tac joint-limit constraints.

\begin{figure*}[!t]
    \centering
    \includegraphics{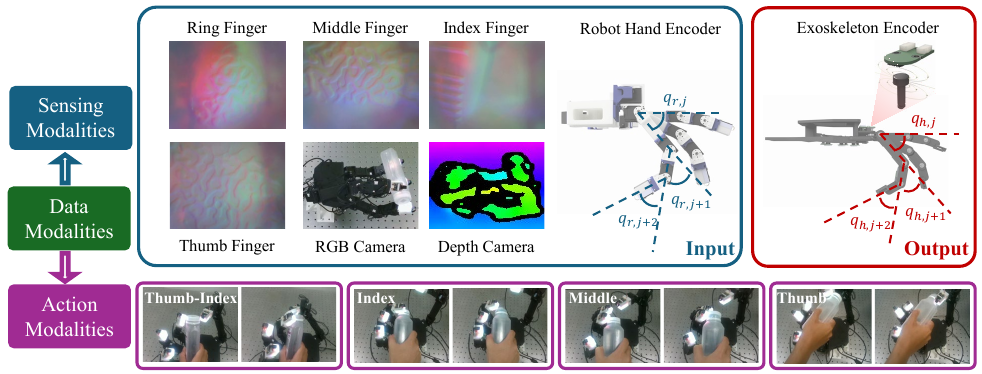}
    \caption{Dataset overview. Synchronized records comprise task-specific visual observations, four fingertip visuotactile streams, the 17-DoF MILE-Tac hand joint state, and the corresponding 17-DoF robot-hand joint-command targets generated in real time from the MILE-exoskeleton measurements through the calibrated affine map. The 196-demonstration bottle-cap training dataset includes external visual observations and multiple finger strategies.}
    \label{fig:supp_dataset}
\end{figure*}

\subsection{Pipeline Timing}
\begin{table}[!t]
    \caption{\rev{Pipeline timing and observed frequencies.}}
    \label{tab:supp_teleop_timing}
    \centering
    \footnotesize
    \setlength{\tabcolsep}{3pt}
    \begin{tabular}{@{}lrrrr@{}}
        \hline
        \rev{Interface} & \rev{P50 (ms)} & \rev{P95 (ms)}
        & \rev{P99 (ms)} & \rev{Frequency (Hz)} \\
        \hline
        \rev{MANUS} & \rev{1.53 / 3.84} & \rev{62.76 / 4.45}
        & \rev{156.46 / 4.61} & \rev{61.37} \\
        \rev{RealSense} & \rev{16.10} & \rev{32.62}
        & \rev{42.75} & \rev{29.93} \\
        \rev{MILE} & \rev{2.39} & \rev{2.54}
        & \rev{2.61} & \rev{417.46} \\
        \hline
    \end{tabular}
\end{table}

\rev{The MANUS entries in Table~\ref{tab:supp_teleop_timing} list the
cross-host and local intervals in that order. The cross-host interval
runs from skeleton publication by MANUS Core on Windows to the Ubuntu
SDK callback, with correction for clock offset. The local interval
runs from skeleton receipt at the Ubuntu bridge to joint-target receipt
by a ROS subscriber. The medians of the two separately measured
intervals sum to 5.37~ms. For the vision timing test, we used a RealSense D435 and
measured from RGB frame retrieval to generation of the filtered joint
target. For MILE, the interval runs from initiation of an angle-read request
to return of a validated joint-angle measurement. It includes serial
request--response communication over a 1~Mbps link, CRC validation,
and angle conversion. The measured angles are directly mapped to
robot-hand joint commands. The Frequency column reports observed event
rates over the respective recording windows: joint-target receptions
at the ROS subscriber for MANUS, filtered joint-target generations for
RealSense, and successful encoder reads for MILE. These are
stage-specific measurements and do not include physical robot-response
latency.}

\subsection{Additional Grasping and Teleoperation Demonstrations}

Fig.~\ref{fig:supp_teleoperation_demonstration} complements the
quantitative task-level comparison in the main manuscript by presenting additional
grasp configurations and continuous manipulation demonstrations.
Panel~(a) shows representative examples spanning enveloping and lateral
grasps, precision pinches, object holding, and dexterous manipulation
with objects of different geometries. Panel~(b) presents continuous
teleoperation sequences for a complex-shaped toy and a basketball. All examples were executed through the MILE exoskeleton--MILE-Tac hand teleoperation interface. Within the selected
17-DoF kinematic structure, the exoskeleton joint measurements were
transmitted as corresponding joint-space commands to the robotic hand.
The examples cover different grasp configurations, contact patterns,
and continuous object-reorientation behaviors, complementing the four
tasks evaluated quantitatively in the main manuscript.

\section{Imitation Learning Protocols and Implementation Details}

\subsection{Robotic-Arm-Mounted Demonstration Collection}

For the robotic-arm-mounted demonstrations, a VIVE Tracker was
attached to the operator side to measure the hand pose. The measured pose
was transformed into the robot-arm coordinate frame and sent to the
Flexiv Rizon 4 as a desired end-effector pose. The arm controller then
generated the corresponding robot-arm motion. In parallel, the MILE
exoskeleton provided joint-space commands to the MILE-Tac hand through
the calibrated correspondence used in the main manuscript. The same arm--hand teleoperation setup was used to collect demonstrations for both the fixed-bottle cap-unscrewing and sequential potato-chip
pick-and-place tasks. The bottle was fixed in the environment for the
fixed-bottle cap-unscrewing task.

\subsection{Camera Configuration}

For the hand-only manipulation tasks, the external RealSense D435 was
mounted on a fixed stand beside and above the hand--object workspace and
oriented toward the manipulation region. For the robotic-arm-mounted
fixed-bottle cap-unscrewing task, a fixed external D435 similarly
provided a third-person view of the hand--object workspace. For
sequential potato-chip pick-and-place, the external D435 was retained,
and an additional RealSense D405 was rigidly mounted on the dorsal side
of the MILE-Tac hand and oriented toward the fingertip manipulation
region to provide an egocentric view.

\subsection{Dataset Overview}

Fig.~\ref{fig:supp_dataset} summarizes the recorded modalities and the
representative bottle-cap training dataset used for policy learning. The upper
part of the figure shows the synchronized sensing and command streams recorded
during demonstration collection, including external visual observations, four
fingertip visuotactile streams, the MILE-Tac hand joint state, and the
exoskeleton-derived robot-hand joint-command targets after the calibrated
affine mapping. The lower part illustrates representative manipulation
strategies in the 196-demonstration bottle-cap dataset, including thumb-only,
index-only, middle-only, and thumb--index cooperative manipulation. These records provide synchronized multimodal demonstrations, while
data processing and synchronization details are described in
Section~VII-D.

\subsection{Data Processing, Synchronization, and Policy Implementation}

During demonstration collection, the four fingertip visuotactile streams were displayed on a monitor as auxiliary visual contact feedback for the operators. Encoder samples were aligned to the visual--tactile streams using nearest-timestamp matching to construct synchronized multimodal datasets. The action field stores the mapped robot-hand joint command sent to the MILE-Tac hand, and the observation state stores the measured MILE-Tac hand joint state.

For tactile policy inputs, the normalized current image $\mathbf{I}_t$ and the no-contact reference image $\mathbf{I}_0$, captured at the beginning of each collection or inference episode, were transformed as $\mathbf{T}_t=(\mathbf{I}_t-\mathbf{I}_0+1)/2$. Because $\mathbf{I}_t-\mathbf{I}_0$ lies in $[-1,1]$, this affine mapping places the signed differences in $[0,1]$, with zero difference represented by $0.5$.

\rev{As shown in Fig.~6 of the main manuscript, ACT-Tac feeds encoded
fingertip tactile tokens into the Transformer alongside visual tokens,
proprioceptive tokens, and the latent query.}

\rev{In DP-Tac, visual and fingertip tactile images are encoded using
separate ViT encoders for the two modalities, with encoder weights
shared across streams within each modality. At each observation step,
the image features and a projected proprioceptive token are fused
through a single self-attention layer with two attention heads.
The fused image features are then concatenated with the original
proprioceptive state. Features from the observation history provide global conditioning
for the one-dimensional U-Net that denoises the action sequence.}

\rev{ACT-Tac predicted a 50-step action chunk. DP-Tac used a 32-step action prediction horizon.
The DP-Tac denoiser was trained to predict noise using a DDPM schedule
with 100 diffusion timesteps. At inference, DP-Tac used DDIM sampling
with 10 denoising steps and extracted 24 action commands from each
predicted sequence. Both policy backbones were trained with a batch
size of 16. For ACT-Tac, the initial learning rates were
$1\times10^{-5}$ for the main network and $3\times10^{-6}$ for the
ViT encoders. The corresponding rates for DP-Tac were
$1\times10^{-4}$ and $1\times10^{-5}$.}

For each task and policy backbone, the paired tactile and non-tactile variants used the same demonstrations, train/validation split, and fixed training schedule. The variants were trained for the same number of epochs with otherwise matched settings, except for the tactile input and associated tactile branch. One model was trained for each variant, and the final checkpoint was used for real-robot evaluation. Table~II of the main manuscript reports real-robot trial results from one trained model per variant.